\documentclass[runningheads]{llncs}
\usepackage{graphicx}

\usepackage{tikz}
\usepackage{comment}
\usepackage{amsmath,amssymb} 
\usepackage{color}

\usepackage{multirow}
\usepackage{subfigure}
\usepackage{caption}
\usepackage{sidecap}
\usepackage{float}
\usepackage[accsupp]{axessibility}  

\begin{document}
\pagestyle{headings}
\mainmatter
\def\ECCVSubNumber{1259}  

\title{Unitail: Detecting, Reading, and Matching in Retail Scene} 

\titlerunning{Unitail: Detecting, Reading, and Matching in Retail Scene}
%
\author{Fangyi Chen\inst{1} \and
Han Zhang\inst{1} \and
Zaiwang Li\inst{2} \and
Jiachen Dou\inst{1} \and
Shentong Mo\inst{1} \and
Hao Chen\inst{1} \and
Yongxin Zhang\inst{3} \and
Uzair Ahmed\inst{1} \and
Chenchen Zhu\inst{1} \and
Marios Savvides\inst{1} 
}
\authorrunning{F. Chen et al.}
%
\institute{Carnegie Mellon University, Pittsburgh PA 15213, USA
\and
University of Pittsburgh, Pittsburgh PA 15213, USA
\and
Tsinghua University, Beijing 100084, China
\\
\email{\{fangyic,hanz3,jiachend,shentonm,haoc3,uzaira,marioss\}@andrew.cmu.edu zal17@pitt.edu yx-zhang20@mails.tsinghua.edu.cn chenchez@alumni.cmu.edu}
}
\maketitle

\begin{abstract}
To make full use of computer vision technology in stores, it is required to consider the actual needs that fit the characteristics of the retail scene. Pursuing this goal, we introduce the United Retail Datasets (Unitail), a large-scale benchmark of basic visual tasks on products that challenges algorithms for detecting, reading, and matching. With 1.8M quadrilateral-shaped instances annotated, the Unitail offers a detection dataset to align product appearance better. Furthermore, it provides a gallery-style OCR dataset containing 1454 product categories, 30k text regions, and 21k transcriptions to enable robust reading on products and motivate enhanced product matching. Besides benchmarking the datasets using various start-of-the-arts, we customize a new detector for product detection and provide a simple OCR-based matching solution that verifies its effectiveness. The Unitail and its evaluation server is publicly available at  \url{https://unitedretail.github.io}.

\end{abstract}

\section{Introduction}
With the rise of deep learning, numerous computer vision algorithms have been developed and have pushed many real-world applications to a satisfactory level. Currently, various visual sensors (fixed cameras, robots, drones, mobile phones, etc.) are deployed in retail stores, enabling advanced computer vision methods in shopping and restocking. Scene Product Recognition (SPR) is the foundation module in most of these frameworks, such as planogram compliance, out-of-stock managing, and automatic check-out.

SPR refers to the automatic detection and recognition of products in complex retail scenes. It comprises steps that first localize products and then recognize them via the localized appearance, analogous to many recognition tasks. However, scene products have their characteristics: \textit{they are densely-packed, low-shot, fine-grained, and widely-categorized}. These innate characteristics result in obvious challenges and will be a continuing problem. Recent datasets in retail scenes follow the typical setting in the common scene to initiate the momentum of research in SPR. For instance, the SKU110k dataset \cite{SKU110k}, which has recently enabled large-scale product detection, is following the MS-COCO \cite{lin2014microsoft} style annotation and evaluation metric. Despite their significant value, the underpaid attention to the SPR's characteristics leads us to the question: 
what is the next advance towards SPR?

Firstly, traditional detection targets poorly comply with the actual needs, causing improper image alignment of the product appearances. Detection targets in common scenes \cite{lin2014microsoft,yang2016wider,voc_dataset,DOTA2021} are usually defined as covering the utmost visible entirety of an object with a minimal rectangle box. This format is inherited by most existing retail datasets \cite{SKU110k,rpc,Grozi-3.2k,Grocery_shelves,locount}. However, because occlusion occurs more frequently between products (the densely-packed characteristic), improper alignments can easily hinder the detection performance. Detectors equipped with Non-Maximum Suppression (NMS) suffer from the overlaps among the axis-aligned rectangular bounding boxes (AABB) and rotated rectangular bounding boxes (RBOX). Moreover, poor alignment leads to inconsistent image registration of the same products, which brings extra difficulties to accurate recognition.   

Secondly, even in the well-aligned cases, products from intra-classes require discriminative features due to their fine-grained characteristic. On the one hand, a slight variation in the product packaging can significantly change the product price, especially for the visually similar but textually different regions such as brand/model, flavour/version, ingredient/material, count/net weight. This requires SPR algorithms to pay attention to the particular text patterns. On the other hand, due to the labelling effort on thousands of categories per store (the widely-categorized characteristic), the available samples per category are scarce (the low-shot characteristic), which degrades the SPR robustness. These two constraints are in conjunction with our empirical observation that visual classifiers could frequently make mistakes when products look similar but vary in text information.

\begin{figure}[t]
    \centering
    \includegraphics[width=\columnwidth]{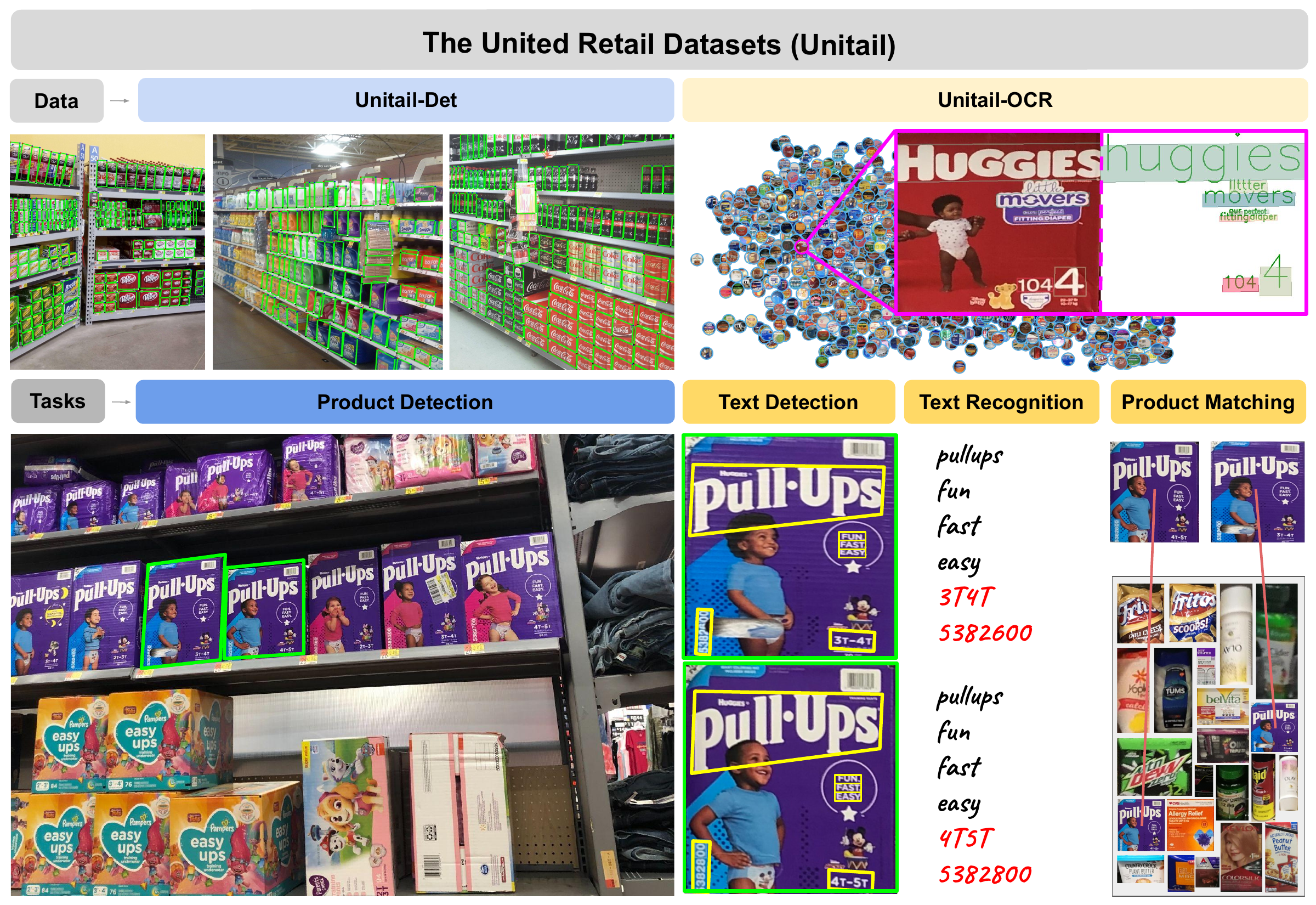}
    \caption{The Unitail is a large-scale benchmark in retail scene that consists of two sub-datasets and supports four basic tasks.}
    \label{fig:overview}
\end{figure}

In this paper, we introduce the United Retail Datasets (Unitail) that responds to these issues. The Unitail is a comprehensive benchmark composed of two datasets: \textit{Unitail-Det} and \textit{Unitail-OCR}, and currently supports four tasks in real-world retail scene: \textit{Product Detection}, \textit{Text Detection}, \textit{Text Recognition}, and \textit{Product Matching}. (Fig.\ref{fig:overview})

Unitail-Det, as one of the largest quadrilateral object detection datasets in terms of instance number and the only existing product dataset in quadrilateral annotations by far, is designed to support well-aligned product detection. Unitail-Det enjoys two key features: 1. Bounding boxes of products are densely annotated in the quadrilateral style that cover the frontal face of products. Practically, Quadrilaterals (QUADs) adequately reflect the shapes and poses of most products regardless of the viewing angles, and efficiently cover the irregular shapes. The frontal faces of products provide distinguishable visual information and keep the appearances consistent. 
2. In order to evaluate the robustness of the detectors across stores, the test set consists of two subsets to support both origin-domain and cross-domain evaluation. While one subset shares the domain with the training set, the other is independently collected from other different stores, with diverse optical sensors, and from various camera perspectives. 

Unitail-OCR (Optical Character Recognition) aims to drive research and applications using visual texts as representations for products. This direction is partially inspired by the customers' behavior: people can glance and recognize ice cream but need to scrutinize the flavor and calories to make a purchase. It is organized into three tasks: text detection, text recognition, and product matching. Product images in Unitail-OCR are selected from the Unitail-Det and benefit from the quadrilateral aligned annotations. Each is equipped with on-product text location and textual contents together with its category. Due to the product's low-shot and widely-categorized characteristics, product recognition is operated by matching within an open-set gallery. To the best of our knowledge, Unitail-OCR is the first dataset to support OCR models' training and evaluation on the retail products, and it is experimentally verified to fill in the domain blank; when evaluated on a wide variety of product texts, models trained on Unitail-OCR outperform those trained on common scene texts \cite{ICDAR15}. It is also the first dataset that enables the exploration of text-based solutions to product matching.


Based on the proposed Unitail, we design two baselines. To detect products, we analyze the limitation of applying generic object detectors in the retail scene and design RetailDet to detect quadrilateral products. To match products using visual texts on 2D space, we encode text features with spatial positional encoding and use Hungarian Algorithm \cite{hungarian_algorithm} that calculates optimal assignment plans between varying text sequences. 

Our contributions are summarized in three folds: \textbf{(1)} we introduce the Unitail, a comprehensive benchmark for well-aligned textually enhanced SPR. \textbf{(2)} We benchmark the tasks of Unitail with various off-the-shelf methods, including \cite{FCENet2021,PANet2019,PSENet2019,DBNet2020,RIDet,xu2019gliding,RSDet,he2017mask,CRNN,sheng2019nrtr,yue2020robustscanner,li2019SAR,SATRN,fang2021ABINet,densebox,ResNet,efficientnetv2,pan2018IBNNet}. \textbf{(3)} we design two baselines for product detection and text-based product matching. 

\section{Related Work}
The retail scene has drawn attention to the computer vision community for an extended period. The early evolution of product datasets \cite{Grozi-120,SOIL-47,Supermarket2010} facilitates reliable training and evaluation and drives research in this challenging field. Recently, large-scale datasets \cite{AliProduct,SKU110k} has enabled deep learning based approaches. Datasets related to SPR can be split into two groups: product detection and product recognition. We address each in this section. 

\subsubsection{Product Detection Datasets}
Detection is the first step in SPR entailing the presence of products that are typically represented by rectangular bounding boxes. The GroZi-120 \cite{Grozi-120} was created using in situ and in vitro data to study the product detection and recognition with 120 grocery products varying in color, size, and shape. The D2S \cite{D2S2018} is an instance segmentation benchmark for sparsely-placed product detection, with 21000 images and 72447 instances. The SKU110k\cite{SKU110k} and the SKU110k-r\cite{SKU110k-r} provide 11762 images with 1.7M on-shelf product instances; they are annotated with axis-aligned bounding boxes and rotated bounding boxes, respectively. 

For the detection of coarsely categorized products, The RPC \cite{rpc} is designed for checkout scenarios containing 0.4M instances and 200 products. The Grocery Shelves dataset \cite{Grocery_shelves} took 354 shelf images and 13k instances, and around 3k instances are noted in 10 brand classes. The Locount \cite{locount} simultaneously considers the detection and counting for groups of products, with an impressive number of 1.9M instances and 140 categories.

Despite the significant values of these datasets, we are still challenged by the availability of optimal bounding boxes. In the proposed Unitail-Det, we aim to shape products into quadrilaterals whose appearances are well aligned. The dataset also provides evaluation targets in the origin-domain and cross-domain, bringing algorithms closer to practical deployment. 

\subsubsection{Product Recognition Datasets}
Multiple product recognition datasets have been built over the past decades. Early ones have limitations in the diversity of categories and amount of images, such as the SOIL-47 \cite{SOIL-47} containing 47 categories, the Supermarket Produce Datasets  \cite{Supermarket2010} with 15 categories, and the Freiburg Groceries \cite{Freiburg_Groceries} covering 25 categories. Recent collections like Products-6K\cite{products-6k} and Products-10K\cite{products-10k} focus on large-scale data, which satisfy the training of deep learning algorithms. 
AliProducts\cite{AliProduct} is crawled from web sources by searching 50K product names, consequently containing 2.5 million noisy images without human annotations. The ABO\cite{ABO_dataset} dataset covers 576 categories that studies the 3D object understanding. The Grozi-3.2K \cite{Grozi-3.2k} contains 80 fine-grained grocery categories and 8350 images for multi-label classification.
To the best of our knowledge, there is a lack of a dataset that encourages leveraging both visual and textual information for product recognition. 

\subsubsection{OCR on Retail Products}
Product recognition by texts is challenging. The lack of datasets obstructs the relevant research on this topic. The most relevant dataset is Products-6K\cite{products-6k} where the Google Cloud Vision API is employed to extract the textual information to enhance products' descriptions. But the texts were not labelled by human annotators, and text location information is missing, so it is infeasible to support any advance to OCR related tasks. 

There are a couple of attempts that use off-the-shelf OCR models for assisted shopping. \cite{assist_shopping} presented a system on which users search products by name and OCR models return the texts on products so that a product list ranked by word histogram
is generated for the users. \cite{mondrianocr} recognize texts and then apply text embedding and language models to extract features for product verification.

\subsubsection{Other Related Datasets}
Many OCR datasets \cite{singh2021textocr,iiit5k,totaltext,ICDAR15,icdar2013,ctw1500,CUTE80,MSRA-TD500,Synth90k,SynthText} exist prior to Unitail-OCR. Typically, an OCR dataset supports text detection and text recognition and so enables text spotting. 
The ICDAR2015\cite{ICDAR15} and CTW1500\cite{ctw1500} are two widely applied benchmarks for training and evaluating OCR models in common scene. The ICDAR2015 has 1,000 training and 500 testing images causally shot indoor and outdoor with word-level annotations for each text. The CTW1500 contains 10k text (3.5k of them are curved boxes) in 1500 images collected from the internet. Compared to them, Unitail-OCR is the first that focuses on product texts and supports object-level matching task at the same time. Product texts are usually artistic words with substantial character distortions, which are hard to be localize and recognize. 

\section{The Unitail}

\subsection{Unitail-Det}
\subsubsection{Image Collection} 
Practically, the industry utilizes a variety of sensors under different conditions for product detection. The resolution and camera angles cover an extensive range by different sensors.
For example, fixed cameras are mounted on the ceiling in most cases, and customers prefer to photograph with mobile devices. The product categories in different stores also span a great range.
With these factors in mind, we collect images from two sources to support origin-domain and cross-domain detection. 
\textit{In the origin domain}, training and testing images are supposed to share the same domain and are taken from similar perspectives in the same stores by the same sensors. As a result, we select the 11,744 images from the prior largest product dataset, SKU110k \cite{SKU110k}, to form the origin domain. 
\textit{In the cross domain}, we collect 500 images in different stores through multiple sensors, covering unseen categories and camera angles.

\begin{figure}[t]
    \begin{minipage}{.48\columnwidth}
    \centering
    \includegraphics[width=\columnwidth]{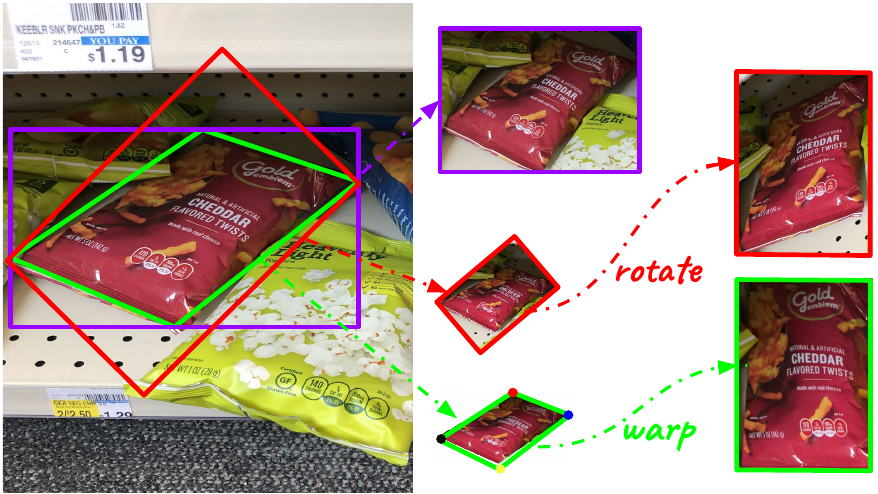}
    \caption{Quadrilateral (in green) is a nature fit to product in real scene, removing more noisy contexts than AABB (in violet) and RBOX (in red)).}
    \label{fig:advance_of_quad}
    \end{minipage}%
    \hfill
    \begin{minipage}{.47\columnwidth}
    \centering
    \includegraphics[width=\columnwidth]{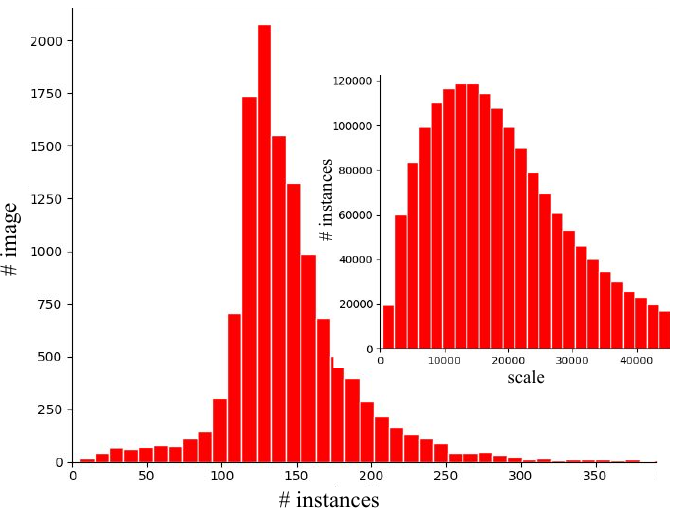}
    \caption{Unitail-Det statistics. Left: instance density. Right: instance scale}
    \label{fig:det_stats}
    \end{minipage}
\end{figure}

\subsubsection{Annotation} We annotate each product with a quadrilateral style bounding box, denoted as QUAD. Fig.\ref{fig:advance_of_quad} is an illustration of its advance. A QUAD refers to 4 points $p_{tl}$, $p_{tr}$, $p_{br}$, $p_{bl}$ with 8 degrees of freedom ($x_{tl}$, $y_{tl}$, $x_{tr}$, $y_{tr}$, $x_{br}$, $y_{br}$, $x_{bl}$, $y_{bl}$). For regular products shaped mainly in cuboid and cylinder, the ($x_{tl}$,$y_{tl}$) is defined as the top-left corners of their frontal faces, and the other points represent the rest 3 corners in clockwise order. For spherical, cones, and other shapes whose corners are difficult to identify, and for irregularly shaped products where so defined quadrilateral box cannot cover the entire frontal face, we first draw the minimum AABB and then adjust the four corners according to the camera perspective. 
As belabored, the frontal face of a product has the most representative information and is also critical for appearance consistency, but we still annotate the side face if the front face is invisible. 

Totally, 1,777,108 QUADs are annotated by 13 well-trained annotators in 3 rounds of verification. The origin-domain is split to training (8,216 images, 1,215,013 QUADs), validation (588 images, 92,128 QUADs), and origin-domain testing set (2,940 images, 432,896 QUADs). The cross-domain supports a testing set (500 images, 37,071 QUADs). Their density and scale are shown in Fig.\ref{fig:det_stats}

\begin{SCfigure}
    \centering
    \includegraphics[width=0.5\columnwidth]{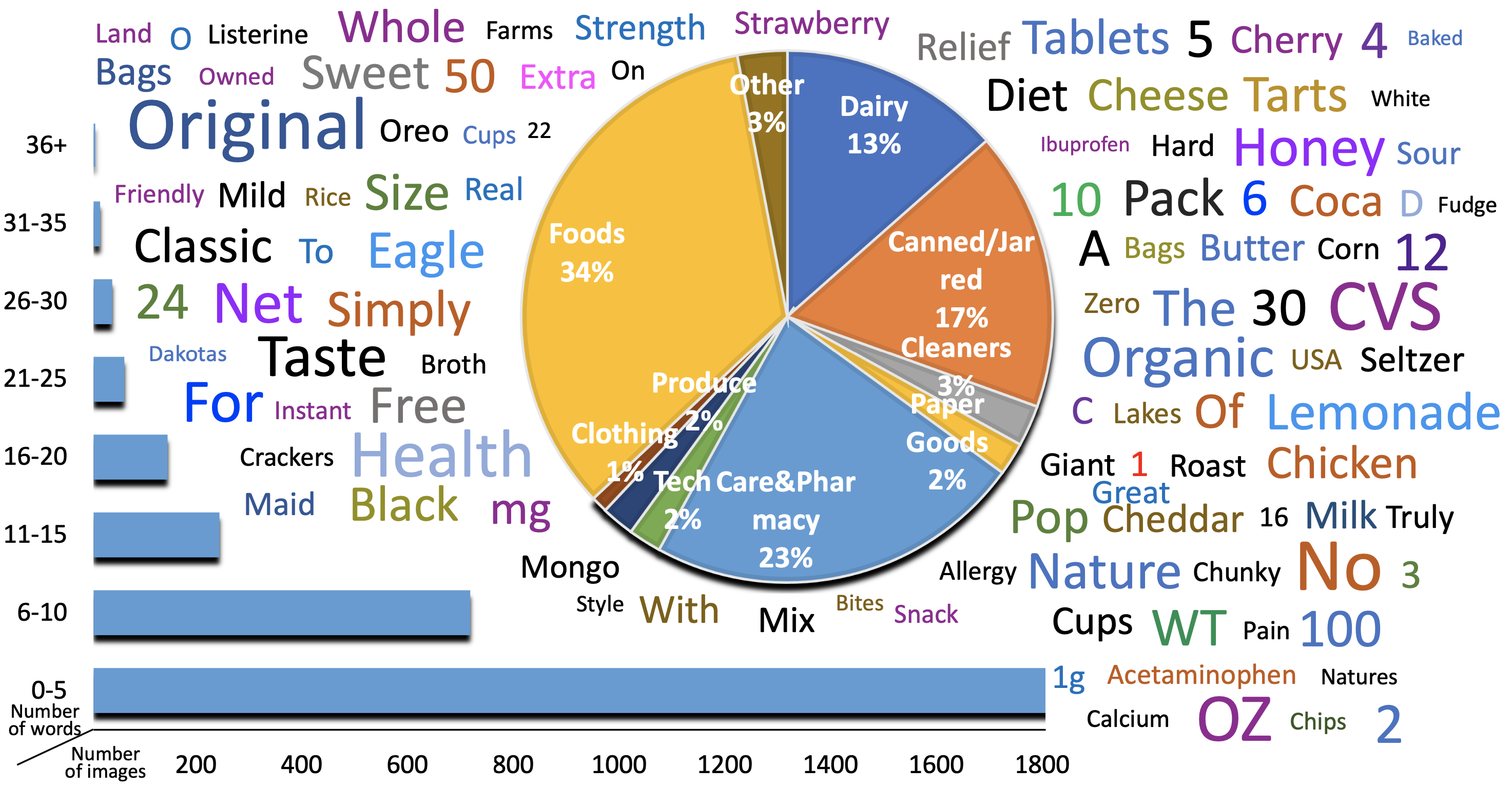}
    \caption{Unitail-OCR statistical graphic. The pie chart reflects sections that source images were collected from. The bar chart is a histogram for the count of words on products. The font size of the words reflects the frequency of occurrence.}
    \label{fig:ocr_stats}
\end{SCfigure}

\subsection{Unitail-OCR}

\subsubsection{Gallery and Testing Suite}
A product gallery setup is a common practice in the retail industry for product matching applications. All known categories are first registered in the gallery. In case of a query product, the matching algorithms find the top ranked category in the gallery. 
The gallery of Unitail-OCR contains 1454 fine-grained and one-shot product categories. Among these products, 10709 text regions and 7565 legible text transcriptions (words) are annotated. This enables the gallery to act as the training source and the matching reference.

The testing suite contains four components: 1. 3012 products labeled with 18972 text regions for text detection. 2. Among the pre-localized text regions, 13416 legible word-level transcriptions for text recognition. 3. 10k product samples from the 1454 categories for general evaluation on product matching. 4. From the 10k products, we select 2.4k fine-grained samples (visually similar for humans) for hard-example evaluation on product matching.

\subsubsection{Image Collection and Annotation} 
Images are gathered from the Unitail-Det cross-domain and cropped and affine transformed according to the quadrilateral bounding boxes to form an upright appearance. We remove the low-quality images with low resolution and high blurriness. Some products kept in the Unitail-OCR might exclude text regions, like those from the produce and clothes departments. We randomly select one sample from each category to form the product gallery, and the remaining is further augmented by randomly adjusting the brightness and cropping for matching purposes. 

We annotate 29681 text regions from 4466 products as quadrilateral text boxes. Fig.\ref{fig:ocr_stats} shows the statistics. The bounding boxes are first classified as \textit{legible} or \textit{illegible}. For the 20981 legible ones, the alphanumeric transcriptions are annotated ignoring letter case and symbols. Numerical values with units are commonly seen on products such as \textit{120mg}, and we regard them as entire words. We also provide a vocabulary that covers all words present. The usage of vocabulary is more practical in our case than in other scenes \cite{ICDAR15}, because the presence of products and texts are usually known in advance by the store owner. 

\subsection{Tasks and Evaluation Metrics}
\subsubsection{Product Detection Task}
The goal is to detect products as quadrilaterals from complex backgrounds. Unitail-Det supports the training and evaluation.

We use the geometric mean of mean average precision (mAP) calculated on the origin-domain test set and cross-domain test set as the primary metric for the product detection, where the mAP is calculated in MS-COCO style\cite{lin2014microsoft}. Compared to arithmetic mean, the geometric mean is more sensitive when the model overfits to origin-domain but gains low performance on the cross-domain. 

\subsubsection{Text Detection Task}
The goal is to detect text regions from pre-localized product images. Unitail-OCR supports the training and evaluation.

We adopt the widely used precision, recall and hmean \cite{ICDAR15,ctw1500} for evaluation. 

\subsubsection{Text Recogniton Task}
The goal is to recognize words over a set of pre-localized text regions. Unitail-OCR supports the training and evaluation.

We adopt the normalized edit distance (NED) \cite{icdar2013} and word-level accuracy for evaluation. The edit distance between two words is defined by the minimum number of characters edited (insert, delete or substitute) required to change one into the other, then it is normalized by the length of the word and averaged on all ground-truths.

\subsubsection{Product Matching Task} 
The goal is to recognize products by matching a set of query samples to the Unitail-OCR gallery. The task is split into two tracks: \textbf{Hard Example Track}, which is evaluated on 2.5k selected hard examples; this track is designed for scenarios in which products are visually similar (for example pharmacy stores). And \textbf{General Track}, which is conducted on all 10k samples. 

We adopt the top-1 accuracy as the evaluation metric.

\section{Two Baselines Designed for The Unitail}

\subsection{A Customized Detector for Product Detection}
Recent studies \cite{zhang2020bridging,tian2019fcos,zhu2020soft,foveabox,Chen2020NCMS} on generic object detection apply prior-art DenseBox-style head \cite{densebox} to multiple feature pyramid levels. The feature pyramid is generated via feature pyramid network (FPN) \cite{lin2016feature} and contains different levels that are gradually down-sampled but semantically enhanced. An anchor-free detection head is then attached to classify each pixel on the feature pyramid and predict axis-aligned bounding boxes (AABB). 

During training, assigning ground-truths to each feature pixels on the feature pyramid plays a key role. \textit{On each pyramid level}, the centerness \cite{tian2019fcos} is widely used. It is an indicator to value how far a pixel locates from the center of a ground-truth: the farther, the more likely it is to predict an inaccurate box, and the lower centerness score it gains. \textit{Across pyramid levels}, various strategies are proposed to determine which level should be assigned, and they are grouped into scale-based and loss-based strategies. The scale-based \cite{lin2016feature,foveabox,ren2015faster,tian2019fcos} assigns ground-truths to different levels in terms of their scales. The larger scale, the higher level is assigned so that the needs of receptive field and resolution of feature maps are balanced. The loss-based like Soft Selection \cite{zhu2020soft} assigns ground-truths by calculating their losses on all levels, and trains an auxiliary network that re-weights the losses. Our design, RetailDet, adopts the DenseBox style architecture but predicts the four corners of quadrilateral by 8-channel regression head. During training, we found the prior assignment strategies unsuitable for quadrilateral products, which is specified below.    


\begin{figure}[t]
    \centering
    \includegraphics[width=1\columnwidth]{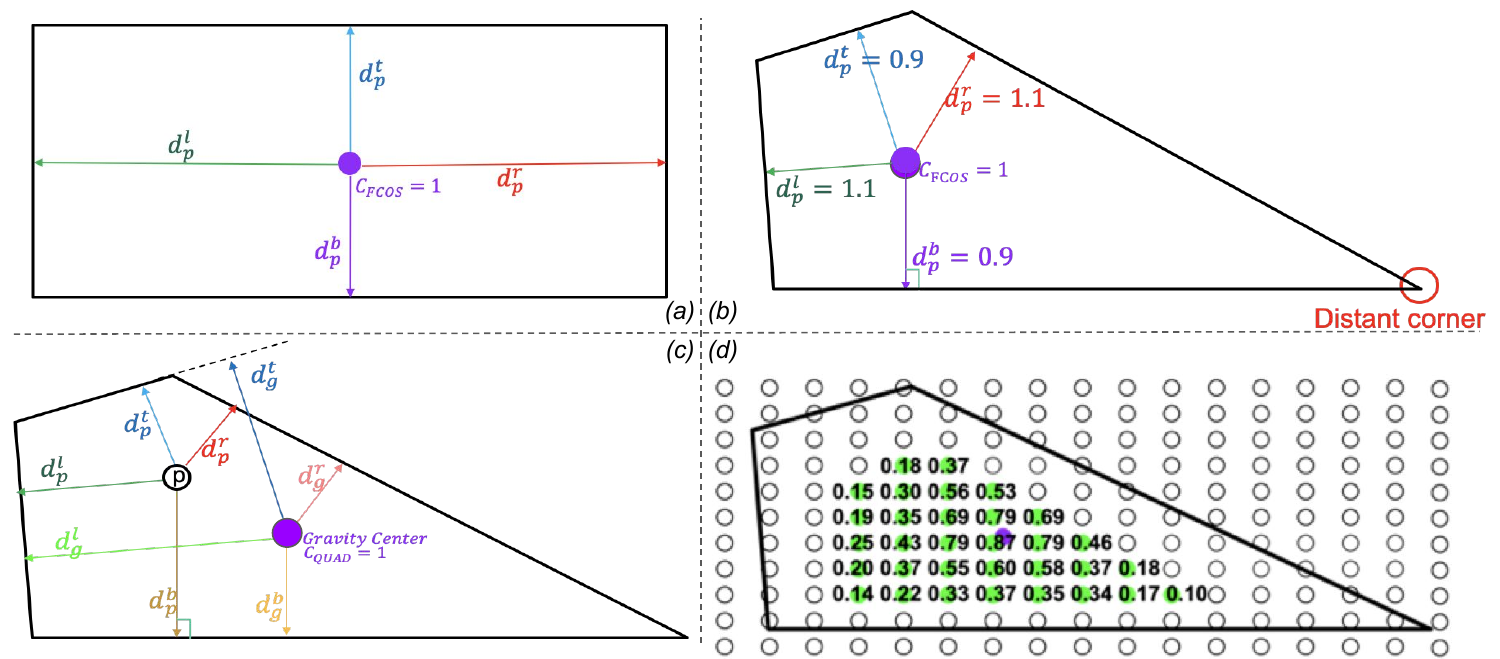}
    \caption{(a): $C_{FCOS}$ on AABB. (b): $C_{FCOS}$ on QUAD. (c) and (d): $C_{QUAD}$.}
    \label{fig:centerness}
\end{figure}

\subsubsection{Centerness}
The previous definition of the centerness \cite{tian2019fcos,zhu2020soft} is shown in Eq.\ref{equ_centerness},

\begin{equation}
C_{FCOS}(p) = [\frac{\min(d_{p}^l,d_{p}^r)}{\max(d_{p}^l, d_{p}^r) } \cdot \frac{\min(d_{p}^t, d_{p}^b)}{ \max(d_{p}^t, d_{p}^b)}]^{0.5}   
\label{equ_centerness}
\end{equation}

\noindent by the Eq.\ref{equ_centerness} and Fig.\ref{fig:centerness}(a), a location $p$ keeps the same distance to the left/right boundaries ($d_{p}^l$ = $d_{p}^r$) and to the top/bottom boundaries ($d_{p}^t$ = $d_{p}^b$) will gain the highest centerness 1, and other pixels gain degraded score by Eq.\ref{equ_centerness}. 

\textbf{Limitation} When adopting the same centerness to quadrilaterals, as shown in Fig.\ref{fig:centerness}(b), the center can be far away from a distant corner, which leads to unbalanced regression difficulty and lack of receptive field from that corner. 

\textbf{Our Solution} We first re-define the \textit{center} as the center of gravity (Fig.\ref{fig:centerness}(c)), because it is the geometric center and represents the mean position of all the points in the shape, which mitigates the unbalanced regression difficulties. We then propose Eq.\ref{equ_centerness_ours} to calculate the quad-centerness for any $p$,

\begin{equation}
C_{QUAD}(p) =
[\frac{\min(d_{p}^l,d_{g}^l)}{\max(d_{}^l, d_{g}^l) } \cdot \frac{\min(d_{p}^r,d_{g}^r)}{\max(d_{p}^r, d_{g}^r) } \cdot
\frac{\min(d_{p}^t,d_{g}^t)}{\max(d_{p}^t, d_{g}^t) } \cdot \frac{\min(d_{p}^b,d_{g}^b)}{\max(d_{p}^b, d_{g}^b) }]^{0.5} 
\label{equ_centerness_ours}
\end{equation}

\noindent where the $d_{g}^{l/r/t/b}$ denotes the distances between the gravity center $g$ and the left/right/top/bottom boundaries. The $d_p^{l/r/t/b}$ denotes the distances between the $p$ and the boundaries. If $p$ locates on the gravity center, its quad-centerness gains the highest value as 1. Otherwise, it is gradually degraded (See Fig.\ref{fig:centerness} (d)). 

It is mentionable that when applied to AABB, Eq.\ref{equ_centerness_ours} is mathematically equivalent to Eq.\ref{equ_centerness}, which is proved in supplementary.

\subsubsection{Soft Selection}
The loss-based Soft Selection in \cite{zhu2020soft} outperforms scale-based strategies on generic objects because it assigns ground-truths to multiple levels and re-weights their losses. This is achieved by calculating losses for each object on all levels and using the losses to train an auxiliary network that predicts the re-weighting factors. 

\textbf{Limitation} Instances per image are numerous in densely-packed retail scene, and Soft Selection is highly inefficient ($5\times$slower) due to the auxiliary network.

\textbf{Our Solution} Can we maintain the merit of Soft Selection while accelerating the assignment? We approach this issue by mimicking the loss re-weighting mechanism of the auxiliary network using scale-based calculation. This is feasible because we find the Soft Selection, in essence, follows scale-based law (detailed in supplementary). Thus, we design Soft Scale (SS) in Eq.\ref{equ_SS_li},\ref{equ_SS_lj},\ref{equ_SS_fli},\ref{equ_SS_flj}. For an arbitrary shaped object $O$ with area $area_O$, SS assigns it to two adjacent levels $l_i$ and $l_j$ by Eq.\ref{equ_SS_li},\ref{equ_SS_lj} and calculates the loss-reweighting factors $F_{li}$, $F_{lj}$ by Eq.\ref{equ_SS_fli},\ref{equ_SS_flj}.

\begin{equation} 
    l_i = \lceil l_{org} + log_2(\sqrt{area_O} / 224) \rceil
\label{equ_SS_li}
\end{equation}
\begin{equation}
    l_j = \lfloor l_{org} + log_2(\sqrt{area_O} / 224) \rfloor
\label{equ_SS_lj}
\end{equation}
\begin{equation}
   F_{l_i} = log_2(\sqrt{area_O} / 224) - \lfloor log_2(\sqrt{area_O} / 224) \rfloor
\label{equ_SS_fli}
\end{equation}
\begin{equation}
    F_{l_j} = 1 - F_{l_i}
\label{equ_SS_flj}
\end{equation}
where 224 is the ImageNet pre-training size.
Objects with exact area $224^2$ is assigned to $l_{org}$, in which case $l_i$ = $l_j$ = $l_{org}$. If an object is with area $223^2$, SS assigns it to $l_{org}$ with $F_{l_{org}} = 0.994$, and also to $(l_{org} - 1)$ with $F_{(l_{org}-1)} = 0.006$. In this work we fix $l_{org}$ to be level 5 of feature pyramid. SS operates rapidly as scale-based strategies and keeps the loss-reweighting like Soft Selection.

\subsection{A Simple Baseline for Product Matching}
\label{product_matching_baseline}
A number of directions can be explored on this new task, while in this paper, we design the simplest solution to verify the motivation: \textit{people glance and recognize the product, and if products looks similar, they further scrutinize the text (if appears) to make decision}. To this end, we first apply a well-trained image classifier that extracts visual features $f_{g_i}^{v}$ from each gallery image $g_{i}$ and feature $f_{p}^{v}$ from query image $p$, and calculate the cosine similarity between each pair $(f_{g_i}^{v}, f_{p}^{v})$ (termed as $sim_i^{v}$). If the highest ranking value $sim_1^{v}$ and the second highest $sim_2^{v}$ are close ($sim_1^{v}$ - $sim_2^{v}$ $\leq$ $t$), we then read on products and calculate the textual similarity (termed as $sim^{t}$) to make decision by Eq.\ref{equ:product_matching_decision},

\begin{equation}
    Decision = \underset{i\in [1,2]}{\operatorname{argmax}}\left[w\cdot sim^{t}(g_{i}, p) + (1-w) \cdot sim^{v}_i\right]
\label{equ:product_matching_decision}
\end{equation}

\noindent where threshold $t$ and coefficient $w$ are tuned on validation set.

\begin{figure}[t]
    \centering
    \includegraphics[width=\columnwidth]{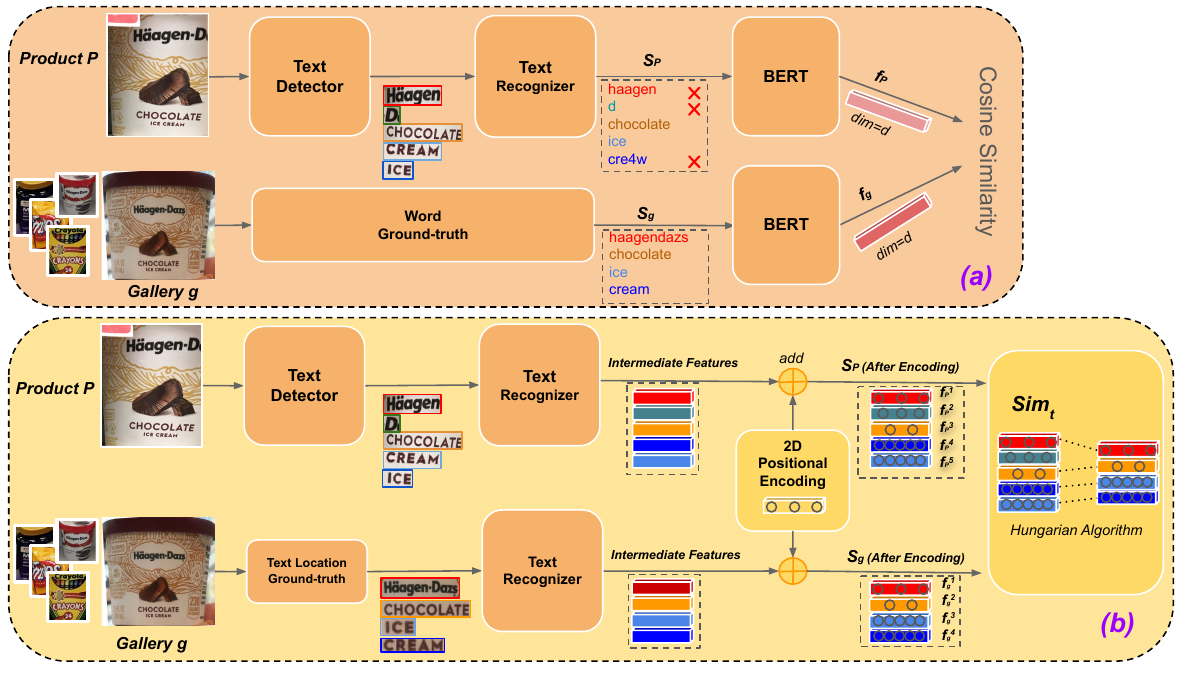}
    \caption{(a) Pipeline with BERT encoded features, (b) Pipeline with positional encoding and Hungarian Algorithm based textual similarity.}
    \label{fig:similarity}
\end{figure}

Our design focuses on how to calculate $sim^{t}$. We denote the on-product texts obtained from ground-truth or OCR prediction as $S=\{s^1,s^2,\ldots, s^N\}$ where $N$ varies. People may propose to utilize sequence-to-one models (like BERT\cite{BERT}) to encode $S$ into a fixed length feature vector $f \in \mathbb{R}^d$. As shown in Fig.\ref{fig:similarity} (a), a text detector is followed by a text recognizer predicting $n=5$ words, and the $5$ words are fed into the BERT  to encode a feature vector $f_p \in \mathbb{R}^d$. For each gallery image $g$, the same process is operated to get a feature vector $f_{g} \in \mathbb{R}^d$, and $sim^{t}(f_p, f_g)$ is calculated by the cosine similarity.

But this design does not perform well because errors from OCR models (especially from text recognizer) are propagated to the BERT causing poor feature encoding. Moreover, the positional information of text boxes is lost in the sequence. So we design a new method in Fig.\ref{fig:similarity} (b). Rather than using the $n$ recognized words, we use the $n$ intermediate feature vectors from the text recognizer to mitigate propagated errors. For example, \textit{CREAM} is confused as \textit{CRE4w}, but the intermediate feature should maintain information on $A$ and $M$, which is more robust than the false prediction. Each feature is then added by a 2D positional encoding\cite{ImageTransformer,DETR} whose calculation is based on the location of the corresponding text. It encodes the spatial information into the feature and it is predefined to keep the same dimension as the intermediate feature. Finally, we get a sequence that contains the $n$ encoded features $f^{1\sim n}$. As shown we get $S_p = \{f^1_p,f^2_p,f^3_p,f^4_p, f^5_p\}$ from a query product and $S_g = \{f^1_g,f^2_g,f^3_g,f^4_g\}$ from a gallery reference. Inspired by the Hungarian Algorithm \cite{hungarian_algorithm}, we design Eq.\ref{equ:hungsimilarity} to directly calculate the similarity between two sequence with varying length:

\begin{equation} 
    sim^{t}(p, g) = sim^{t}(S_p, S_g) = \underset{X}{\operatorname{max}}\sum_{i=1}^{n}\sum_{j=1}^m (\frac{f_p^i \cdot f_g^j}{|f_p^i|\cdot |f_g^j|} X_{ij})
\label{equ:hungsimilarity}
\end{equation}

\noindent where the $X$ is a $n \times m$ boolean matrix where $\sum_{j}X_{ij} = 1, \sum_{i}X_{ij} = 1$. Eq.\ref{equ:hungsimilarity} maximizes the summation of cosine similarities from assigned feature pairs, and the assignment is optimized by $X$.

\section{Experiments}
\subsection{Benchmarking The Unitail}
The implementation is detailed in the supplementary materials.
\subsubsection{Quadrilateral Product Detection} 
Along with the proposed RetailDet, we build baselines by borrowing the existing detectors capable of detecting quadrilaterals, mainly from the textual and aerial scenes\cite{DOTA2021}. These methods can be grouped into segmentation based methods \cite{FCENet2021,PANet2019,PSENet2019,DBNet2020,he2017mask} and regression based methods \cite{RIDet,xu2019gliding,RSDet}. Segmentation methods consider QUADs as per-pixel classification masks for each region of interest, while regression methods directly regress the bounding boxes. 

Table \ref{tab:detection_benchmark} shows their performances. Overall, regression based methods outperform segmentation based methods because most products can be well aligned by quadrilaterals, and the learning on segmentation mask involves extra difficulties. The RetailDet outperforms other detectors by a large margin. All detectors achieve degraded results in the cross-domain as opposed to the origin-domain, confirming that domain shift exists among stores. 

\begin{table}
\centering
\caption{Benchmarking product detection on the Unitail. All methods are trained and tested under same setting. g-mAP is the geometric mean of mAPs. 
}
\begin{tabular}{ll|l|c|ccc|ccc}
\hline \hline
 &&&& \multicolumn{3}{c|}{Origin-Domain} & \multicolumn{3}{c}{Cross-Domain}\\
\#&Method                    & Backbone &g-mAP     & mAP  & AP50 & AP75   & mAP  & AP50 & AP75 \\ \hline
1&FCENet \cite{FCENet2021}   & ResNet50 &32.0      & 36.8 & 76.0 & 31.2   & 27.9 & 60.1 & 22.6 \\
2&PANet \cite{PANet2019}     & ResNet50 &35.0      & 40.5 & 72.8 & 41.9   & 30.3 & 53.3 & 31.6 \\
3&PSENet \cite{PSENet2019}   & ResNet50 &39.4      & 45.3 & 77.5 & 49.5   & 34.4 & 58.7 & 36.9 \\
4&DBNet \cite{DBNet2020}     & ResNet50 &45.3      & 51.0 & 86.8 & 55.4   & 40.3 & 71.6 & 42.7 \\
5&RIDet \cite{RIDet}         & ResNet50 &45.7      & 51.2 & 82.9 & 58.5   & 40.8 & 70.3 & 43.2 \\
6&Gliding Vertex\cite{xu2019gliding} & ResNet50 &46.0      & 52.3 & 89.0 & 56.9   & 40.5 & 76.7 & 38.6 \\
7&RSDet \cite{RSDet}         & ResNet50 &46.1      & 51.4 & 83.6 & 58.8   & 41.4 & 71.1 & 44.4 \\
8&Mask-RCNN \cite{he2017mask}& ResNet50 &52.4      & 57.3 & 91.6 & 66.0   & 48.0 & 77.9 & 53.2 \\ 
9&RetailDet (ours)    & ResNet50 &54.7      & 58.7 & 91.6 & 68.4   & 50.9 & 80.6 & 56.7 \\
10&RetailDet          & ResNet101&57.1      & 60.3 & 92.8 & 70.6   & 54.1 & 83.5 & 60.6 \\
\hline
\end{tabular}
\label{tab:detection_benchmark}
\end{table}

\subsubsection{Text Detection $\&$ Text Recognition} We benchmark the tasks of text detection and text recognition in Table \ref{tab:textdetection} and Table \ref{tab:textrecog}, respectively. For each of the listed algorithms, it is trained under two setting: one is following a common setting that trained on SynthText\cite{SynthText} and ICDAR2015(IC15)\cite{ICDAR15} for text detection and on Synth90K\cite{Synth90k} and SynthText for text recognition, another is trained/finetuned on the Unitail. As shown all algorithms achieve better performance if trained on the Unitail, this verifies that texts in the retail domain are better handled by the proposed dataset. 

\begin{SCtable}[]
\centering
\setlength\tabcolsep{3pt}
\caption{Benchmarking text detection on Unitail. P and R stand for Precision and Recall, respectively. hmean is the harmonic mean of precision and recall. }
\begin{tabular}{l|l|ccc}
\hline \hline
Method         & Training Set        & R       & P       & hmean    \\ \hline
DBNet\cite{DBNet2020}          & SynthText+IC15      & 0.541   & 0.866   & 0.666    \\
DBNet          & Unitail             & 0.773   & 0.871   & 0.819    \\
FCENet\cite{FCENet2021}         & SynthText+IC15      & 0.420   & 0.745   & 0.538    \\
FCENet         & Unitail             & 0.795   & 0.857   & 0.825    \\
PSENet\cite{PSENet2019}         & SynthText+IC15      & 0.421   & 0.750   & 0.539    \\
PSENet         & Unitail             & 0.705   & 0.789   & 0.744    \\
\hline
\end{tabular}
\label{tab:textdetection}
\end{SCtable}

\begin{SCtable}[]
\centering
\setlength\tabcolsep{1.2pt}
\caption{Benchmarking text recognition on Unitail. S90k: Synth90k. ST: SynthText. PW: methods use public available weights. NED: Normalized Edit Distance, the lower the better. Acc: word top-1 Accuracy, the higher the better.}
\begin{tabular}{l|l|c|ll}
\hline \hline
Method      & Training Set     &  PW            & NED     & Acc(\%)   \\ \hline
CRNN\cite{CRNN}        & S90k+ST            & $\checkmark$   & 0.36    & 40.0     \\
CRNN        & S90k+ST+Unitail    &                & 0.25    & 51.4    \\
NRTR\cite{sheng2019nrtr}        & S90k+ST            & $\checkmark$   & 0.28    & 55.7     \\
NRTR        & S90k+ST+Unitail    &                & 0.16    & 69.4    \\
RobustScanner\cite{yue2020robustscanner}        & S90k+ST            & $\checkmark$   & 0.25    & 56.3     \\
RobustScanner        & S90k+ST+Unitail    &                & 0.18    & 65.9    \\
SAR\cite{li2019SAR}        & S90k+ST            & $\checkmark$   & 0.25    & 56.2     \\
SAR        & S90k+ST+Unitail    &                & 0.18    & 66.5    \\
SATRN\cite{SATRN}        & S90k+ST            & $\checkmark$   & 0.23    & 62.7     \\
SATRN        & S90k+ST+Unitail    &                & 0.13    & 74.9    \\
ABINet\cite{fang2021ABINet}        & S90k+ST            & $\checkmark$   & 0.17    & 69.2     \\
ABINet        & S90k+ST+Unitail    &                & 0.11    & 77.2    \\
\hline
\end{tabular}
\label{tab:textrecog}
\end{SCtable}

\subsubsection{Product Matching}
The product matching results are shown in Table \ref{tab:match_hard_example}. With just texts across all 1454 categories, the method in Fig.\ref{fig:similarity}(b) reaches 31.71\% and 47.81\% on the Hard Example Track and the General Track, respectively. The result is convincing that only textual information is a strong representation for the product, but this new direction clearly requires further exploration.

Moreover, the textual information improves the regular visual classifier using the method proposed in Sec.\ref{product_matching_baseline}. In the hard example track, the improvement of textual information is significant (+$1.76\sim 2.75\%$) since the similar-looking products which are hard for regular classifier could be easier distinguished by texts. In the general track, the improvement drops (+$0.56\sim 0.94\%$) due to the limited ability of the textual matching algorithm. 

\begin{table}[t]
    \begin{minipage}{0.48\columnwidth}
    \centering
    \caption{Benchmarking on the product matching task. }
    \begin{tabular}{l|l}
        \hline \hline
        Method                                       & Acc (\%)  \\ \hline
        Hard Example:                                         \\ 
        Only text (Fig.\ref{fig:similarity} (b))    & 31.71             \\
        EfficientNetV2  \cite{efficientnetv2}                         & 56.49  \\
        EfficientNetV2+Text                         & 59.24 (+2.75)   \\
        ResNet101      \cite{ResNet,pan2018IBNNet}                            & 58.37      \\
        ResNet101+Text                             & 60.52 (+2.15)      \\
        General:  \\
        Only Text(Fig.\ref{fig:similarity} (a))     & 30.37           \\
        Only Text(Fig.\ref{fig:similarity} (b))     & 47.81 (+17.44)           \\
        EfficientNetV2                      & 83.81         \\
        EfficientNetV2+Text                         & 84.62 (+0.81) \\
        ResNet101                          & 85.03              \\
        ResNet101+Text                             & 86.19 (+1.16)      \\
        \hline
    \end{tabular}
    
    \label{tab:match_hard_example}
    \end{minipage}
    \hfill
    \begin{minipage}{0.48\columnwidth}
    \centering
    
    \caption{Results on SKU110k. RetailDet++ is an enhanced variant where a box refinement module is added (see supplementary for details)}
    \begin{tabular}{l|c}
    \hline \hline
    Method          & mAP    \\ \hline
    RetinaNet+EM\cite{SKU110k}   &  49.2  \\
    FSAF    \cite{Zhu_2019_CVPR}         &  56.2  \\
    CenterNet  \cite{duan2019centernet} & 55.8    \\  
    ATSS    \cite{zhang2020bridging}         & 57.4    \\
    FCOS    \cite{tian2019fcos}        & 54.4    \\
    SAPD    \cite{zhu2020soft}        & 55.7    \\
    Faster-RCNN+FPN     \cite{ren2015faster}     &54.0  \\
    Reppoints       \cite{reppoints}   &55.6   \\ 
    Cascade-RCNN+Rong \cite{rong2021solution}  &58.7    \\ \hline
    RetailDet++ (Ours)         &\textbf{59.0}   \\
    \hline
    \end{tabular}
    \label{table_sku110k}
    \end{minipage}
    
\end{table}

\subsection{Discussion}
\subsubsection{RetailDet} Table \ref{table_sku110k} shows that the RetailDet achieves state-of-the-art results on product detection benchmark SKU110k \cite{SKU110k}. The SKU110k and the Unitail-Det share the same training images but with different annotations (QUAD vs AABB), and the origin-domain mAP on the SKU110k is 59.0 and 61.3 on the Unitail-Det (not included in Table.\ref{tab:detection_benchmark}). The difference is mainly for the reason that QUAD is a natural fit for products as a smoother regression target, and it avoids densely-packed overlaps which makes post-processing easier. 
\subsubsection{Difficulty of The Text Based Product Matching} Since the proposed matching pipeline is not end-to-end trained, errors are accumulated by the text detector (0.773 recall and 0.871 precision for DBNet) and text recognizer (0.772 for ABINet). Some products in the dataset do not contain texts (2\%), and many words are partially or fully illegible. This requires further study on topics such as: attention mechanism on text regions, end-to-end framework for text spotting based product recognition, and one-shot learning based sequence matching. 

\section{Conclusions}
In this work, we introduce the United Retail Datasets (Unitail), a large-scale benchmark aims at supporting well-aligned textually enhanced scene product recognition. It involves quadrilateral product instances, on-product texts, product matching gallery, and testing suite. We also design two baselines that take advantages of the Unitail and provide comprehensive benchmark experiments on various state-of-the-art methods.

%
%
\bibliographystyle{splncs04}
\bibliography{egbib}

\clearpage

\appendix

\section{Datasets Comparison}

\begin{table}[H]
\centering
\begin{tabular}{l|cccccccc}
\hline \hline
Dataset         & $\#$Image & $\#$Instance  & Box Type &$\#$Category & Det & Rec & Text & \\ \hline
Grozi-3.2k      & 9,030     & 11,585        & AABB     & 80          & \checkmark                \\
Grocery Shelves & 354       & 13,000        & AABB     & 10          & \checkmark                \\
SKU110k         & 11,748    & 1,730,996     & AABB     & 1           & \checkmark                \\
SKU110k-r       & 11,748    & 1,731,762     & RBOX     & 1           & \checkmark                \\
Locount         & 50,394    & 1,905,317     & AABB     & 140         & \checkmark                \\
RPC             & 30,000    & 367,935       & AABB     & 200         & \checkmark  & \checkmark  \\
Grozi-120       & 11,870    & -             & -        & 120         &             & \checkmark  \\  
SOIL-47         & 987       & -             & -        & 47          &             & \checkmark  \\  
SuperMarket     & 2,633     & -             & -        & 15          &             & \checkmark  \\
Freiburg        & 5,021     & -             & -        & 25          &             & \checkmark  \\
Product10K      & 150,000   & -             & -        & 10k         &             & \checkmark  \\
Unitail(ours)   & 12,244    & 1,777,108     & QUAD     & 1454        & \checkmark & \checkmark & \checkmark \\ \hline
\end{tabular}
\caption{Comparison of related benchmarks.}
\label{table:datasetcomparison}
\end{table}

\section{Annotation}
\subsection{Unitail-Det}
\subsubsection{Annotation Method}
We consider quadrilaterals as proper fits to products. The bounding box is common for localizing objects and reflects their shape in detection tasks. Axis-aligned rectangles are popular because they satisfy the minimum requirement for learning targets with minimum labeling efforts. Annotations with more accurate localization and appearance alignment are needed in this task. While segmentation masks are another level of accuracy for annotating scene objects, they are not cost-friendly, and the direct regression on bounding boxes is easier for densely-packed products than the segmentation methods verified in the benchmark. 

\begin{figure}[t]
    \centering
    \includegraphics[width=\columnwidth]{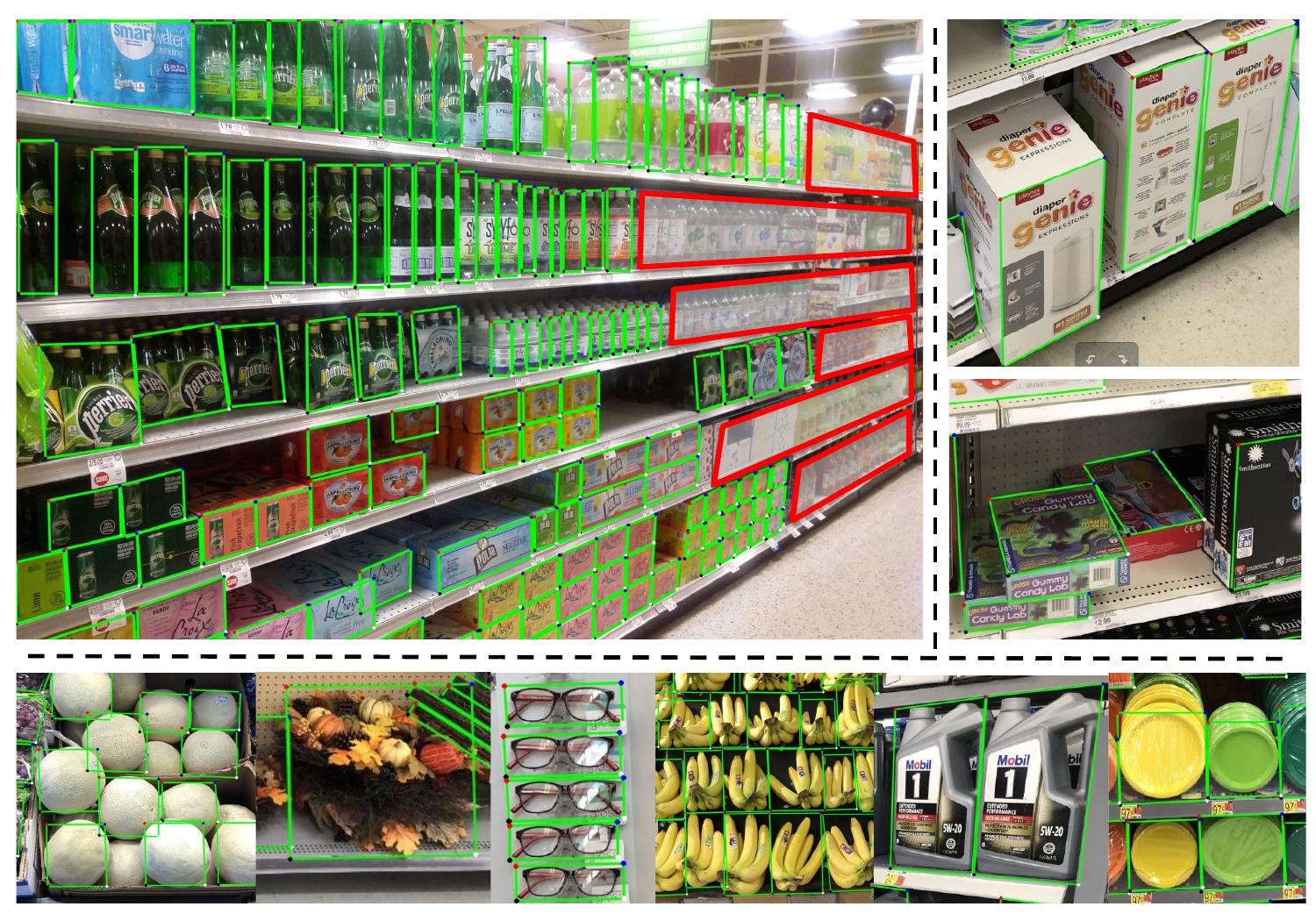}
    \caption{Annotation examples of the Unitail-Det. Top-left: small products are ignored by masks (red bounded regions). Top-right: frontal faces represent products. Bottom: quadrilaterals on irregular-shaped products.}
    \label{fig:det_anno_examples}
\end{figure}

To ensure the quality of annotation, annotators follow a strictly defined guide. Illustrated in Fig.\ref{fig:det_anno_examples}, products localized far away from the camera and with a size less than $8 \times 8$ pixels are not treated as positive. Instead, we annotate ignoring masks covering the distant regions for these products. Cuboid and cylinder (boxes, cans, bottles), as the majority in stores, enjoy normal quadrilaterals defined in the paper; spherical and cones whose corners are difficult to identify, along with irregularly shaped products and distorted bags are expected to be affine-transformed back into an upright position. In those cases, we first draw the minimum AABB to cover them and then adjust the four corners according to the camera perspective. Only the visible part is annotated if another product or tag blocks one product. Many products have multiple faces observable, and only the frontal face of the products is practically annotated. Labelme \cite{wada2018labelme} is applied as an annotation tool.

Although many existing methods \cite{RIDet,RSDet,EAST} re-order the regression targets (four corner points) to favour the loss convergence regardless of the original order in ground-truths, we define the first corner point $(x1, y1)$ as the top-left corner.

\subsubsection{Statistics}
\begin{figure}[t]
    \centering
    \includegraphics[width=\columnwidth]{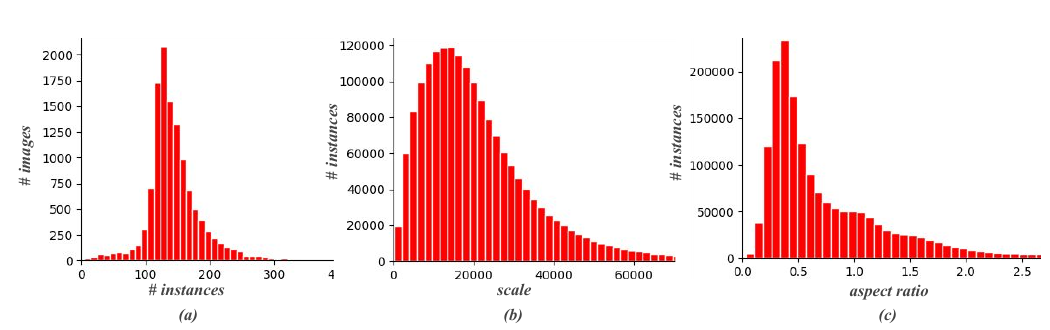}
    \caption{Histograms. (a) Instance density. (b) Instance scales. (c) Instance aspect ratio}
    \label{fig:sup_fig_det_stats}
\end{figure}

Fig.\ref{fig:sup_fig_det_stats} (a) illustrates the number of QUADs in each image. The mean and standard deviation is 145 and 46, respectively. An image in QuadRetail may contain at least 5 QUADs and up to 744 QUADs. Despite the density, the overlap among QUADs is not severe due to the annotation standard. 

Fig.\ref{fig:sup_fig_det_stats} (b) illustrates the scales of QUADs. The average scale is 22393.7 ($149.6^2$) over the QuadRetail. The minimum and maximum are $17^2$ and $1938^2$, respectively. The average image width is 2466.8, and the average height is 3288.1. 

The aspect ratio of a rectangle is commonly defined as $\frac{w}{h}$. We measure the aspect ratio of QUAD as in Eq.\ref{equ:AR}

\begin{equation}
ratio = \sqrt{\frac{t \cdot b}{l \cdot r}}   
\label{equ:AR}
\end{equation}

\noindent where $t$, $b$, $l$, $r$ are lengths of top, bottom, left, and right edge, respectively. Fig.\ref{fig:sup_fig_det_stats} (c) illustrates the aspect ratio. Most ratios are around 0.3 - 0.6, which is in line with the practical observation. There are also QUADs with extreme AR ($\textless 0.05$) and ($\textgreater 38$).

The interior angles of any convex QUAD add up to 360 degrees of arc. The standard deviation of these angles ($std_a$) is a qualified reflection of the QUAD shape. For rectangles, $std_a = 0$. For a extreme QUAD looks close to a line segment, $std_a=90$. In the origin-domain of QuadRetail, the $std_a=6.24$. In the cross-domain, the $std_a=12.73$. This means the images from the cross-domain are from taken from tougher angles.

\subsection{Unitail-OCR}
Unitail-OCR consists of a gallery and a testing suite to support text detection, text recognition, and product recognition.
\subsubsection{Annotation Method}
We start by cropping products from the Unitail-Det cross-domain. Since the crops are in quadrilaterals, they are further transformed to form upright appearances. Products with scales larger than $15 \times 15$ are further selected. The categorization is organized by two hierarchical stages. In the first stage, ten supercategories (food, dairy, paper goods, canned, produce, clothing, Technology, Pharmacy, Care, other) are defined to classify the products coarsely. In the second stage, we first apply a strong pre-trained model trained with a bag of tricks to group the product into 6k clusters and correct them by human annotators. During the correction, we do not accept blurred products that are hard to identify. Many fine-grained categories rely on textual information, and the blurred ones that 
are difficult to read are filtered out. After the categorization, we label each product with word-level text boxes, following the annotation method in ICDAR2015 \cite{ICDAR15}. We annotate 29681 text regions from 4466 products as quadrilateral text boxes. The bounding boxes are classified as legible or illegible. The alphanumeric transcriptions $0\sim 9, a\sim z$ are annotated for the legible ones. A vocabulary covers all words in the gallery and the testing suite are attached with the dataset.

\section{Baseline Implementation}
\subsection{Off-the-shelf Algorithms}
We conduct experiments using multiple codebases including mmdetection \cite{mmdetection} for FCOS, SAPD, ATSS; mmocr \cite{mmocr2021} for DBNet, FCENet, PSENet, CRNN, NRTR, RobustScanner, SAR, SATRN, ABINet; AlphaRotate \cite{yang2021alpharotate} for RIDet and RSDet; maskrcnn benchmark\cite{massa2018maskrcnn_benchmark} for maskrcnn and Gliding Vertex; timm \cite{rw2019timm} for efficientnet, ResNet, and ResNet-IBN. To be more specific, models for product detection tasks are trained on 4 NVIDIA Titan RTX GPUs with two images per GPU. For fair comparison, the training schedule is 12 epochs with an initial learning rate of 0.01 divided by 10 at the 9th and the 11th epoch. Unless otherwise specified, the input images are scaled to 1200 and randomly horizontally flipped without any other augmentation. The Convex Hull and Shoelace Formula are implemented in CUDA to calculate the exact IOU of QUADs. Up to 400 detections per image are allowed to evaluate. For text detection models, we respect their optimized training setting for ICDAR2015 based on their officially released paper or code, but change the input image size to $1333 \times 800$. For text recognition, we finetune the publicly available weights on Unitial-OCR for 10 epochs.

\subsection{RetailDet and RetailDet++}
\subsubsection{Base Network}
Our design of base network applies prior-art DenseBox-style head \cite{densebox} to multiple feature pyramid levels. The feature pyramid is generated via feature pyramid network (FPN) \cite{lin2016feature} which utilizes a deep convolutional network as the backbone. As an image is fed into the backbone, several feature maps are extracted to compose the initial feature pyramid. This work adopts the ResNet family as the backbone, and the extracted feature maps are from $C_3$ to $C_5$. The feature maps after FPN are denoted as $P_3$, $P_4$, $P_5$. An anchor-free detection head is attached then. The head contains two branches. One is a binary classification branch to predict a heatmap for product/background. Another is a regression branch to predict the offset from the pixel location to the four corner points of the QUAD. Each branch consists of 3 stacks of convolutional layers followed by another $c$ channel convolutional layer, where $c$ equals to 1 for the classification branch and 8 for the regression branch.

\subsubsection{Corner Refinement Module}
RetailDet++ is the RetailDet enhanced with Corner Refinement Module (CRM) and deeper backbone. Here, we introduce the CRM. For each predicted QUAD from the RetailDet, we get the locations of its four corners and center. Then we apply the bilinear interpolation to extract feature of 5 points (4 corners, one center) from the feature map generated by the 3rd stacked convolution in the regression branch. These features are concatenated and fed into a $1\times1$ convolutional layer to predict the offsets between ground-truth and the former predictions. The same operation and convolution are also inserted in the classification branch to predict retail/background as a 2nd-stage classification. During testing, we combine the regression results from the two stages but only use the classification result from the first stage. CRM shares the spirits with Faster-RCNN\cite{ren2015faster}, BorderDet\cite{BorderDet} and Reppoints \cite{reppoints}, but we find that the 5 points as mentioned above are enough for quadrilateral products, and the 2nd-stage classification helps training though not involved in testing.

\subsubsection{Losses}
During training, we first shrink QUADs by a ratio $\alpha = 0.3$  according to the gravity centers. If one feature pixel locates inside the shrunk QUAD, the pixel is considered responsible for learning the ground-truth. 
We utilize $focal~ loss$ \cite{lin2017focal} for classification and $SmoothL_1~loss$ for regression, and we reweight both losses by the production of quad-centerness and level reweighting factor $F$. The total loss is the summation of the classification and regression losses. If two-stage, additional $focal~ loss$ and $L_1~loss$ for CRM are added to the total loss.

\section{Discussion}
\subsection{Proof: Eq.2 Is Equivalent to Eq.2 on Rectangles}
When QUAD is specialized to Rectangles, in Eq.2,  $d_g^l=d_g^r$, $d_{p}^l+d_{p}^r=2d_g^l$, so if $d_p^l \textless d_g^l$, then $d_p^r \textgreater d_g^l = d_g^r$; if $d_p^l \textgreater d_g^l$, then $d_p^r \textless d_g^l = d_g^r$. Thus, $\frac{\min(d_{p}^l,d_{g}^l)}{\max(d_{p}^l, d_{g}^l) } \cdot \frac{\min(d_{p}^r,d_{g}^r)}{\max(d_{p}^r, d_{g}^r) } = \frac{\min(d_{p}^l,d_{p}^r)}{d_{g}^l } \cdot \frac{d_{g}^r}{\max(d_{p}^l, d_{p}^r) } =  \frac{\min(d_{p}^l,d_{p}^r)}{\max(d_{p}^l, d_{p}^r) }$, similarly to $d^t$ and $d^b$,  then, Eq.2 is mathematically equivalent to Eq.1.

\subsection{Analysis on Soft Selection}
Soft Selection is a loss-based strategy, where training losses of ground-truths indicate their pyramid level. It first assigns each object to all pyramid levels $P_3$, $P_4$, $P_5$ and calculates $loss_l$ for each level $P_l$. $l$=3,4,5. Then, the level that produces the minimal loss is converted to a one-hot vector, i.e., ($1$,$0$,$0$) if the minimal loss is from $P_3$; and ($0$,$1$,$0$) if it is from $P_4$, and so on. The vector is used as the ground-truth to train an auxiliary network that simultaneously predicts a vector ($F_3$, $F_4$, $F_5$). Each element $F_l$ is a down-weighting factor for $loss_l$. The final loss of each object is $\sum_l (F_l \cdot loss_l)$. 

By Soft Selection, the minimal loss from level $l$ indicates that the auxiliary network is trained to generate a relatively larger $F_l$, but we find the loss not independent of scales. On the contrary, object scale inherently determines which level will produce the minimal loss. We claim the reason as follows. First, when assigning objects (e.g. object $A$ with size $8\times8$ and $B$ with size $16\times16$) to pyramid, their regression targets (denoted as $T_A$, $T_B$) are normalized by the level stride. Specifically, on a lower level like $P_3$, the target is divided by stride 8, while on a higher level like $P_4$, the target is divided by 16, and so on. Therefore, when assigning $A$ to $P_3$ and $P_4$, $T_A$ is $1\times1$ and $0.5\times0.5$, respectively; when assigning $B$, $T_B$ is $2\times2$ and $1\times1$, respectively. Note that all levels share the detection head. Apparently, the combination of $T_A=1\times1$ and $T_B = 1\times1$ leads to the smallest regression difficulty for the regression head. Naturally, it produces minimal regression losses, which means the smaller object is assigned to a lower level. Second, since $A$ has a smaller scale, it requires more local fine-grained information beneficial for classification, which is more available from high-resolution lower levels. In comparison, $B$ has a larger scale and needs a larger receptive field, which is more available from higher levels. Therefore, the "loss-based" Soft Selection, in essence, follows the scale-based law. 

Nevertheless, why does Soft Selection outperforms scale-based strategies? We credit the improvement to its loss-reweighting mechanism. This mechanism involves multiple levels during training and reweights the loss in terms of the regression and classification difficulties, making optimization easier. Since the pyramid is discrete, if an object scale falls into the gap of two adjacent levels, both levels' difficulties will be similar. The auxiliary network has opportunities to learn to predict proper $F_l$ for both levels. The analysis motivates us to abandon the auxiliary network and design Soft Scale (SS).

\section{Additional Results}

\subsection{RetailDet}
\subsubsection{On SKU110k-R}
The result of RetailDet on SKU110k-R is compared with other methods in Table.\ref{table_sku110kr}. RetailDet outperforms the state-of-the-art detectors CenterNet\cite{xingyi2019centernet}, DRN \cite{SKU110k-r} and CFA \cite{Guo_2021_CFA} on SKU110k-r where products are in RBOX style. Following CFA, we use multi-scale training.
\begin{table}[t]
\centering
\begin{tabular}{l|ccc}
\hline \hline
Method           & mAP    & AP75    & AR300 \\ \hline
YoloV3-Rotate\cite{redmon2018yolov3}    & 49.1   & 51.1    & 58.2  \\
CenterNet-4point\cite{xingyi2019centernet} & 34.3   & 19.6    & 42.2  \\
CenterNet\cite{xingyi2019centernet}        & 54.7   & 61.1    & 62.2  \\
DRN\cite{SKU110k-r}              & 55.9   & 63.1    & 63.3  \\ 
CFA\cite{Guo_2021_CFA}              & 57.0   & 63.5    & 63.9  \\ 
RetailDet  & \textbf{57.5}   & \textbf{65.5}    & \textbf{64.3}  \\ 
\hline
\end{tabular}
\caption{Detection performance on SKU110k-r.}
\label{table_sku110kr}
\end{table}

\subsubsection{Ablation Study} Table.\ref{table_ablation_retaildet} shows the improvement of each component brings to RetailDet. The Quad-Centerness (QC), Soft-Scale (SS) and 
Corner Refinement Module (CRM) gradually improve the mAP by 2.1, 1.0, 2.4 in the origin domain and 1.8, 0.6, 3.2 in the cross domain. And the improvement is consistent under different IOU thresholds.

\begin{table}[t]
\centering
\begin{tabular}{cccc|ccc}
\hline \hline
Base      &QC & SS & CRM & mAP  & AP50 & AP75 \\ \hline
$\checkmark$    &   &    &          & 58.3 & 87.0 & 69.5  \\
$\checkmark$ &$\checkmark$&&       & 60.4 & 89.3 & 71.5  \\
$\checkmark$ &$\checkmark$&$\checkmark$&&61.4&90.3&72.6 \\
$\checkmark$ &$\checkmark$&$\checkmark$&$\checkmark$&63.8& 91.2 & 75.8 \\
\hline
\end{tabular}
\caption{Ablation study on the Unitail-Det val set. QC: Quad-Centerness. SS: Soft Scale. CRM: Corner Refinement Module.}
\label{table_ablation_retaildet}
\end{table}

\subsection{Inference Speed}
We report the inference speed of the key models in Table \ref{table:Inferencetime}
\begin{table}[]
\centering
\begin{tabular}{l|cccc}
\hline \hline
Methods         & Backbone &Task                  & FPS   \\ \hline
RetailDet       & ResNet101 & Product Detection   & 6.3    \\
DBNet           & ResNet50 & Text Detection       & 27.6    \\
ABINet          & ResNet-ABI & Text Recognition     & 41.4   \\
Visual+Text*       & -   & Product Matching          &  65.1   \\
 \hline
\end{tabular}
\caption{Speed tested on single 2080Ti. * pipeline is accelarated without losing accuracy by applying only text models on visually low-confidence products.}
\label{table:Inferencetime}
\end{table}

\subsection{Qualitative Results}
\begin{figure*}[t]
\centering
\includegraphics[width=\columnwidth]{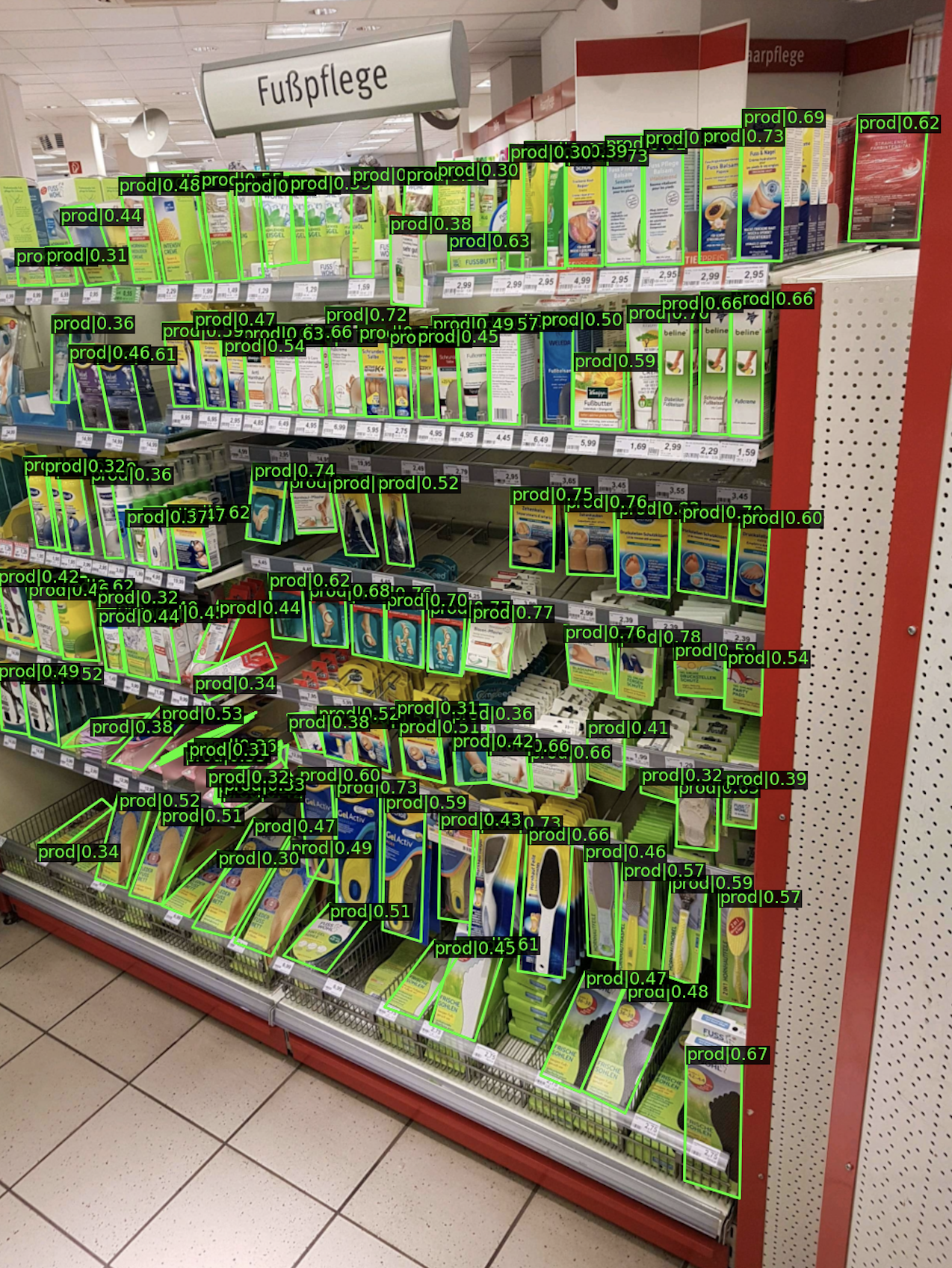}
\caption{Visualization of high difficulty detection result from the RetailDet.}
\label{fig_visual_det_hard}
\end{figure*}

\begin{figure*}[t]
\centering
\includegraphics[width=\columnwidth]{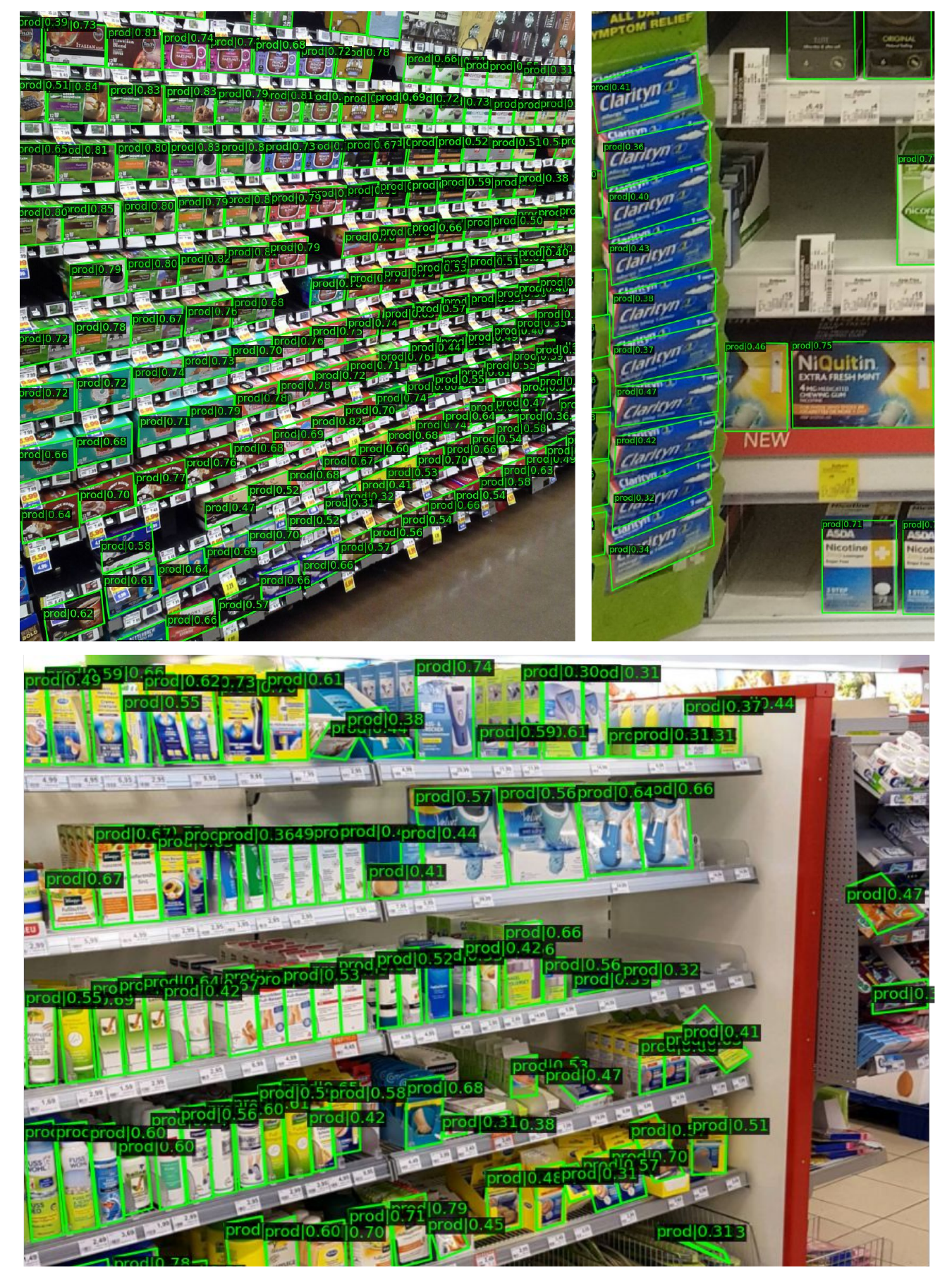}
\caption{Visualization of medium difficulty detection result from the RetailDet.}
\label{fig_visual_det_mid}
\end{figure*}

\begin{figure*}[t]
\centering
\includegraphics[width=\columnwidth]{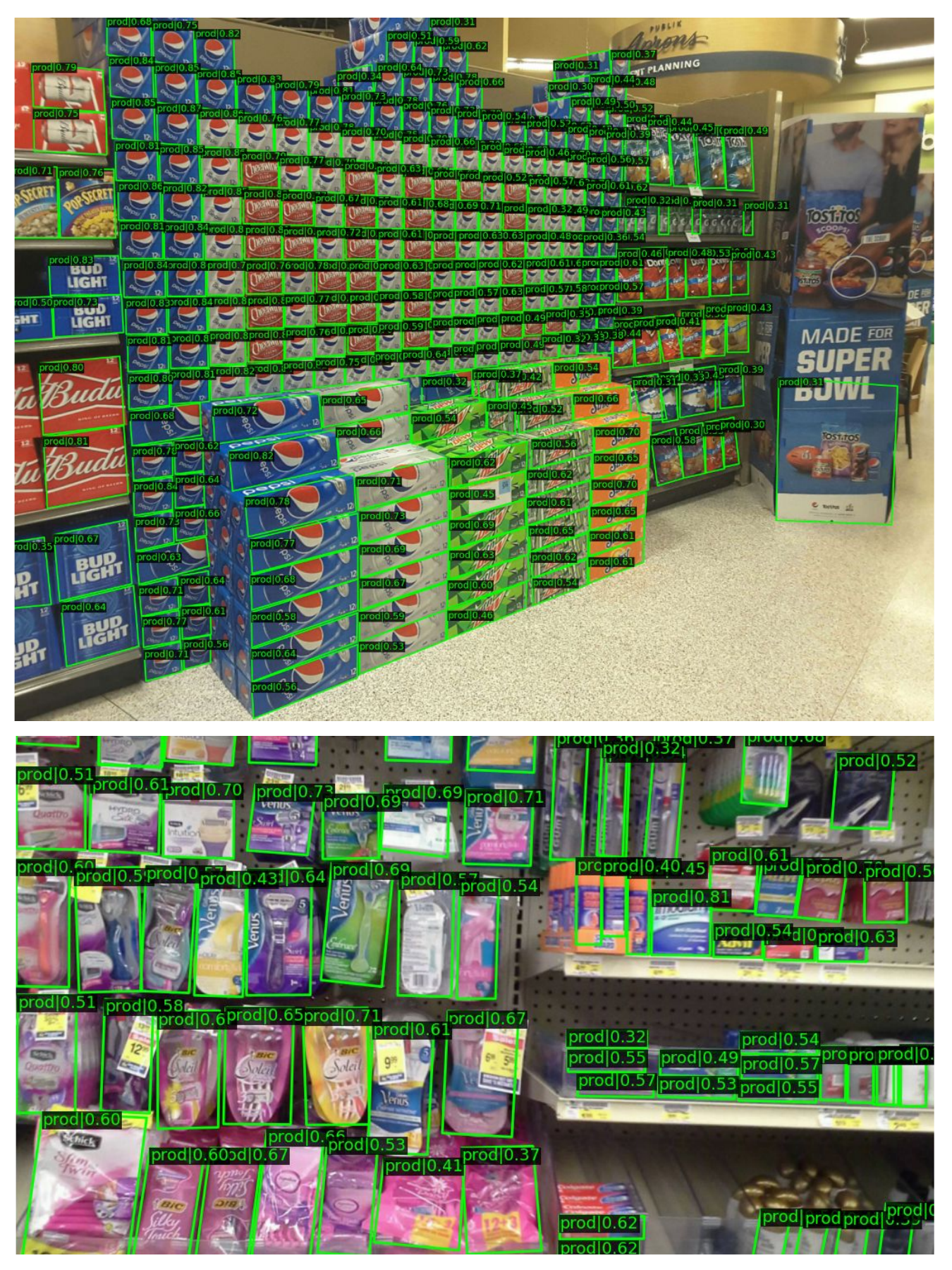}
\caption{Visualization of low difficulty detection result from the RetailDet.}
\label{fig_visual_det_low}
\end{figure*}

We visualize the product detection results in Fig.\ref{fig_visual_det_hard}, Fig.\ref{fig_visual_det_mid}, and Fig.\ref{fig_visual_det_low}. The detector is our RetailDet two-stage variant with ResNext101. The average testing speed is 2.8FPS. We only visualize the QUADs with confident scores higher than 0.3. 

\begin{figure*}[t]
\centering
\includegraphics[width=\columnwidth]{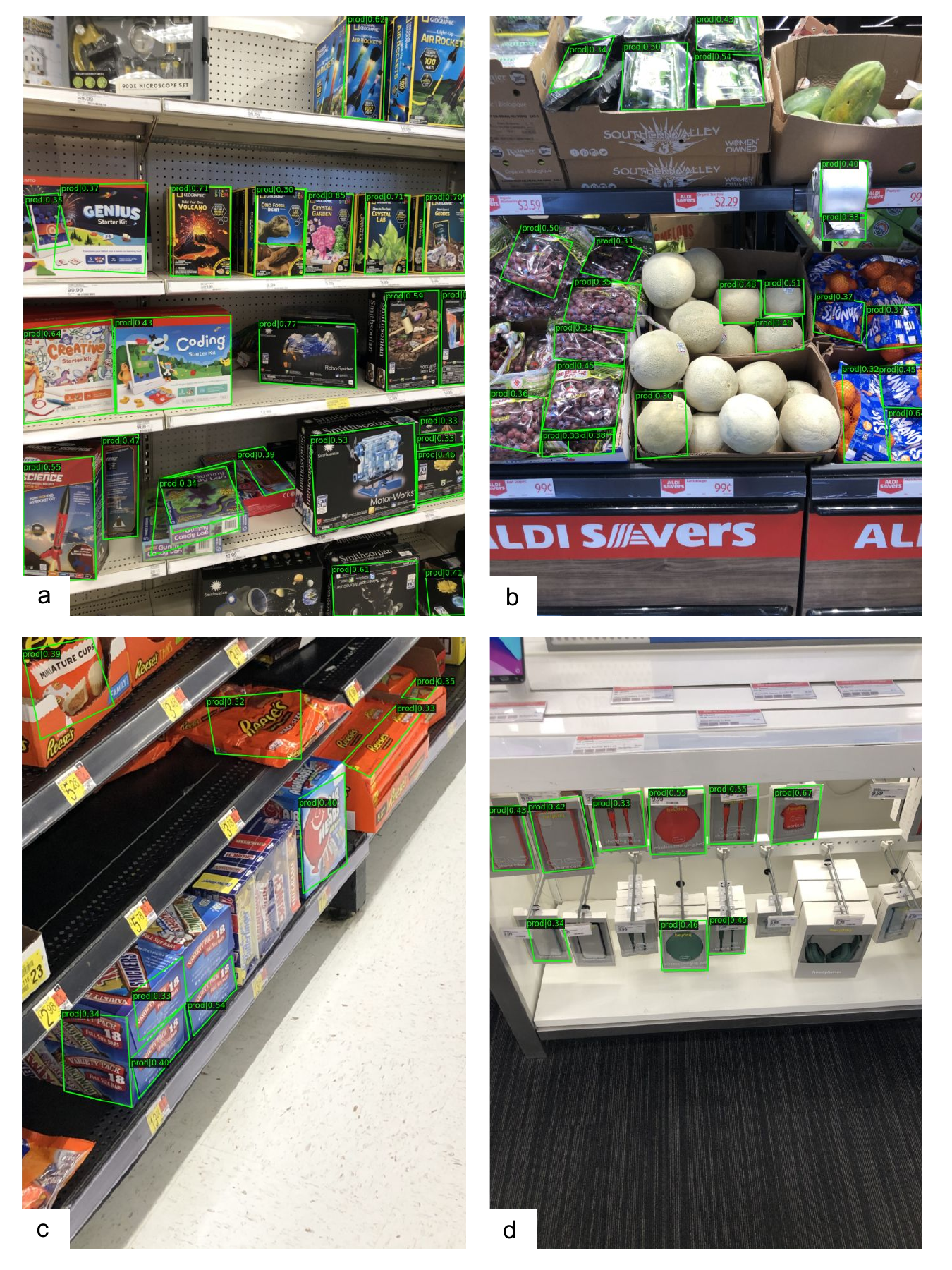}
\caption{Visualization of failure cases in cross domain. (a)(b) Unseen categories. (c)(d) Tough shooting angles.}
\label{fig_visual_failure}
\end{figure*}

We show some failure cases in the cross domain in  Fig.\ref{fig_visual_failure}. (a)(b) Unseen categories. (c)(d) Tough camera angles.

\begin{figure*}[t]
\centering
\includegraphics[width=\columnwidth]{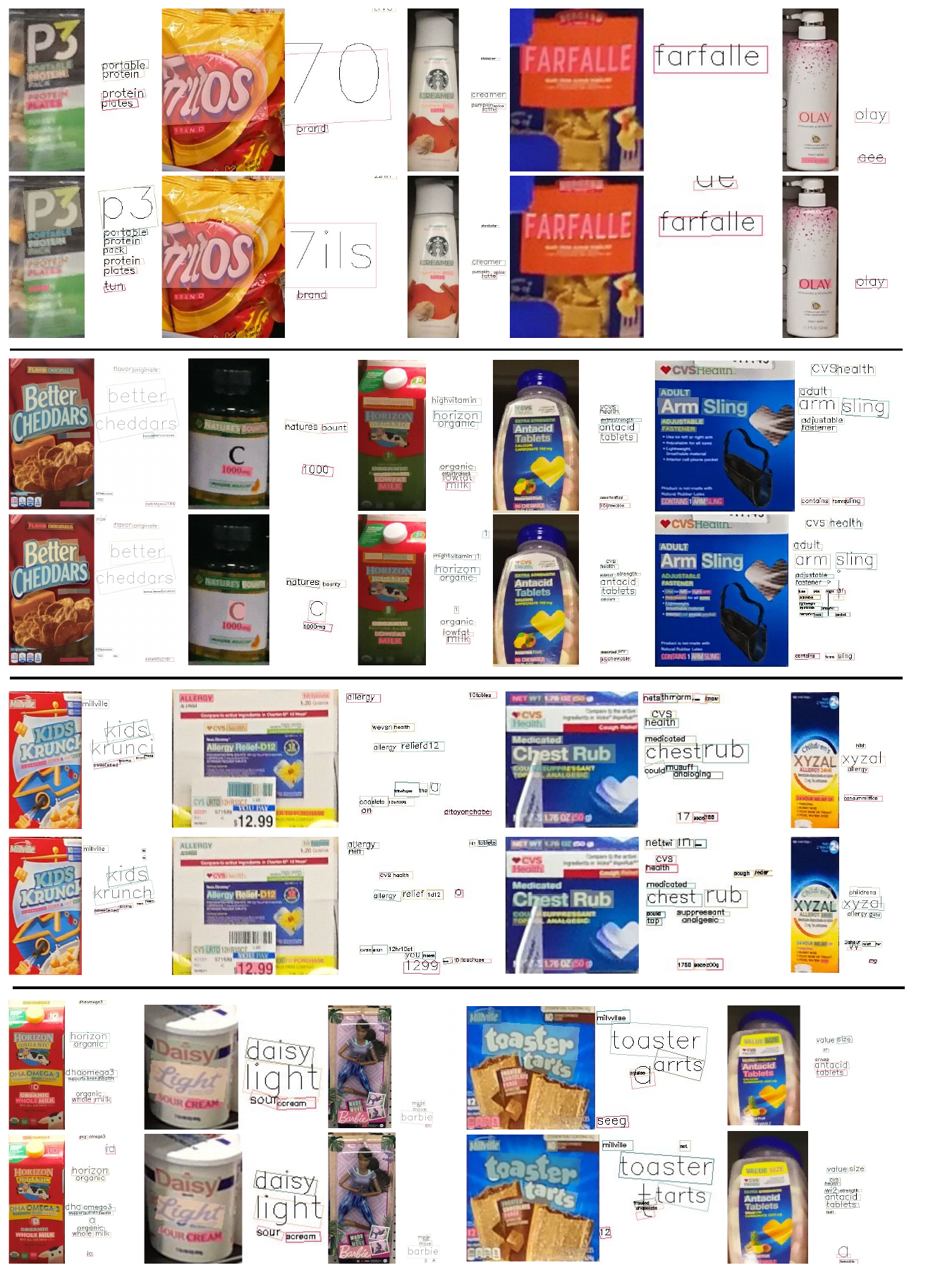}
\caption{Visualization of OCR results. Upper ones of each section are from models trained on ICDAR2015, and lower ones are on Unitail-OCR.}
\label{fig_resocr}
\end{figure*}

We show the OCR results from DBNet and ABINet in Fig.\ref{fig_resocr}.

\end{document}